\pdfoutput=1

\documentclass[11pt]{article}

\usepackage[final]{ACL2023}

\usepackage{times}
\usepackage{latexsym}

\usepackage{graphicx}
\usepackage{float}
\usepackage{wrapfig}
\usepackage{colortbl}
\usepackage{subfigure}
\usepackage{diagbox} 
\usepackage{bm}
\usepackage{times}
\usepackage{latexsym}
\usepackage{listings}
\usepackage{xcolor}
\usepackage{makecell}

\usepackage[T1]{fontenc}

\usepackage[utf8]{inputenc}

\usepackage{microtype}

\usepackage{inconsolata}

\title{TencentPretrain: A Scalable and Flexible Toolkit for Pre-training Models of Different Modalities}

\author{Zhe Zhao\textsuperscript{1*}, Yudong Li\textsuperscript{2}, Cheng Hou\textsuperscript{1}, Jing Zhao\textsuperscript{1}, Rong Tian\textsuperscript{1}, Weijie Liu\textsuperscript{1}, Yiren Chen\textsuperscript{1},\\ 
        {\bf Ningyuan Sun\textsuperscript{1}, Haoyan Liu\textsuperscript{1}, Weiquan Mao\textsuperscript{1}, Han Guo\textsuperscript{1}, Weigang Guo\textsuperscript{1}, Taiqiang Wu\textsuperscript{1},} \\
        {\bf Tao Zhu\textsuperscript{1}, Wenhang Shi\textsuperscript{3}, Chen Chen\textsuperscript{1}, Shan Huang\textsuperscript{1}, Sihong Chen\textsuperscript{1}, Liqun Liu\textsuperscript{1}, Feifei Li\textsuperscript{1},} \\
        {\bf Xiaoshuai Chen\textsuperscript{1}, Xingwu Sun\textsuperscript{1}, Zhanhui Kang\textsuperscript{1}, Xiaoyong Du\textsuperscript{3}, Linlin Shen\textsuperscript{2}, Kimmo Yan\textsuperscript{1}} \\
        \textsuperscript{1} Tencent AI Lab \\
        \textsuperscript{2} School of Computer Science and Software Engineering, Shenzhen University \\
        \textsuperscript{3} School of Information and DEKE, MOE, Renmin University of China}

\begin{document}
\sloppy
\maketitle
\begin{abstract}
Recently, the success of pre-training in text domain has been fully extended to vision, audio, and cross-modal scenarios. The proposed pre-training models of different modalities are showing a rising trend of homogeneity in their model structures, which brings the opportunity to implement different pre-training models within a uniform framework. In this paper, we present TencentPretrain, a toolkit supporting pre-training models of different modalities. The core feature of TencentPretrain is the modular design. The toolkit uniformly divides pre-training models into 5 components: \emph{\textbf{embedding}}, \emph{\textbf{encoder}}, \emph{\textbf{target embedding}}, \emph{\textbf{decoder}}, and \emph{\textbf{target}}. As almost all of common modules are provided in each component, users can choose the desired modules from different components to build a complete pre-training model. The modular design enables users to efficiently reproduce existing pre-training models or build brand-new one. We test the toolkit on text, vision, and audio benchmarks and show that it can match the performance of the original implementations. 
\end{abstract}

\begin{table*}[!htbp]\label{models}
    \centering
    \begin{footnotesize}
    \resizebox{1.00\textwidth}{!}{

    \begin{tabular}{|c|c||c|c|c|c|c|}

    \hline
    \textbf{Pre-training model}&\textbf{Modality}&\textbf{Embedding}&\textbf{Encoder}&\textbf{\thead{Target \\ embedding}}&\textbf{Decoder}&\textbf{Target}\\
    \hline
    ELMo \scriptsize{\cite{peters2018deep}}&Text&word&bi-lstm&-&-&bilm\\
    \hline
    Infersent \scriptsize{\cite{conneau2017supervised}}&Text&word&gru&-&-&cls\\
    \hline
    CoVe \scriptsize{\cite{mccann2017learned}}&Text&word&lstm&word&lstm&lm\\
    \hline
    BERT \scriptsize{\cite{devlin2019bert}}&Text&word, pos, seg&transformer&-&-&mlm, sp\\
    \hline
    GPT-2 \scriptsize{\cite{radford2019language}}&Text&word, pos&transformer&-&-&lm\\
    \hline
    T5 \scriptsize{\cite{raffel2020exploring}}&Text&word&transformer&word&transformer&lm\\
    \hline
    ViT \scriptsize{\cite{dosovitskiy2020image}}&Vision&patch, pos&transformer&-&-&cls\\
    \hline
    BEiT \scriptsize{\cite{bao2021beit}}&Vision&patch, pos&transformer&-&-&mlm\\
    \hline
    S2T \scriptsize{\cite{wang2020fairseq}}&Audio&speech, pos&transformer&word, pos&transformer&lm\\
    \hline
    ViLT \scriptsize{\cite{kim2021vilt}}&Text-vision&word\_patch, pos, seg&transformer&-&-&mlm, cls\\
    \hline
    \end{tabular}
    }
        
\end{footnotesize}
\caption{\small{Typical pre-training models and their modules. The above models use different variants of transformer. Due to the page limit, we do not list the details of transformer module in encoder component. In addition, abbreviations are used in embedding and target columns. pos and seg respectively stand for position and segment embeddings. bilm, cls, lm, mlm, sp respectively stand for bi-directional language model, classification, language model, masked language model, sentence prediction.}}
\end{table*}

\section{Introduction}

\let\thefootnote\relax\footnotetext{*Corresponding Author \\ E-mail: nlpzhezhao@tencent.com}

Pre-training on large-scale data and then fine-tuning on downstream tasks has become a paradigm for text, vision, and audio tasks \cite{devlin2019bert,bao2021beit,baevski2020wav2vec}. In addition to the similarity in the pipeline paradigm, these pre-training models as well have close model structures: On one hand, most of them consist of the following components, \emph{\textbf{embedding}}, \emph{\textbf{encoder}}, \emph{\textbf{target embedding}}, \emph{\textbf{decoder}}, and \emph{\textbf{target}}, on the other hand, many modules in above components are shared among models of different modalities. For example, the transformer module (in encoder component) \cite{vaswani2017attention}, which is successful in the field of text, is increasingly being applied to the vision and audio modalities. \cite{dosovitskiy2020image,gulati2020conformer}. Table 1 lists the commonly used pre-training models and their modules.

The trend towards homogeneity in pre-training models is becoming more apparent, which makes it possible to integrate them into a uniform framework. A representative work in this direction is Huggingface Transformers \cite{wolf2020transformers}, which exploits a non-modular design mode. For each pre-training model in Huggingface Transformers, several separate classes are created, and the code is not refactored with additional abstractions. Users can develop their pre-training models independently which is useful to collaborative development in the community. However, in this design mode, users need to implement the model from scratch when adding a new pre-training model, requiring considerable code work. In addition, with the increased number of pre-training models, the number of classes and lines of code also increases linearly. Codes with the same function may be written many times, which degrades the readability and maintainability of the project.

In response to shortcomings of non-modular design mode, we introduce TencentPretrain, a modular toolkit specially designed for pre-training models of different modalities. As shown in Figure 1, TencentPretrain has five components, namely \emph{\textbf{embedding}}, \emph{\textbf{encoder}}, \emph{\textbf{target embedding}}, \emph{\textbf{decoder}}, and \emph{\textbf{target}}. Among them, \emph{\textbf{target embedding}} and \emph{\textbf{decoder}} components are optional, since the targets of many pre-training models do not involve sequence decoding \cite{zhang2020pegasus,lewis2020bart}. TencentPretrain is hierarchical modular designed with two degrees of freedom. At component level, users are free to combine modules within a component, for example, combining multiple modules in \emph{\textbf{target}} component to perform multi-task pre-training \cite{lan2019albert,sun2020ernie}. At the model level, users can combine modules from different components to constitute a complete pre-training model.

Modularity in design makes TencentPretrain scalable with the increasing number of newly proposed pre-training models. Users are allowed to reuse existing modules with little efforts, avoiding repeated implementation of core functions. At the same time, TencentPretrain provides a robust and clear interface among different components. It brings flexibility, allowing users to build custom model structures through a configuration file without any code work.

TencentPretrain is implemented with PyTorch \cite{paszke2019pytorch}, and it supports distributed training and DeepSpeed optimization library \cite{rasley2020deepspeed}. TencentPretrain is fully connected with Huggingface Transformers, providing comprehensive conversion scripts of pre-training models between the two frameworks. Users can switch between the two frameworks at low cost. TencentPretrain is tested on text, vision, and audio benchmarks and is able to reproduce the results of SOTA pre-training models. The TencentPretrain toolkit is publicly available at \url{https://github.com/Tencent/TencentPretrain}.

\begin{figure*}
\centering
\setlength{\belowcaptionskip}{-0.5cm}
\includegraphics[width=6.1in]{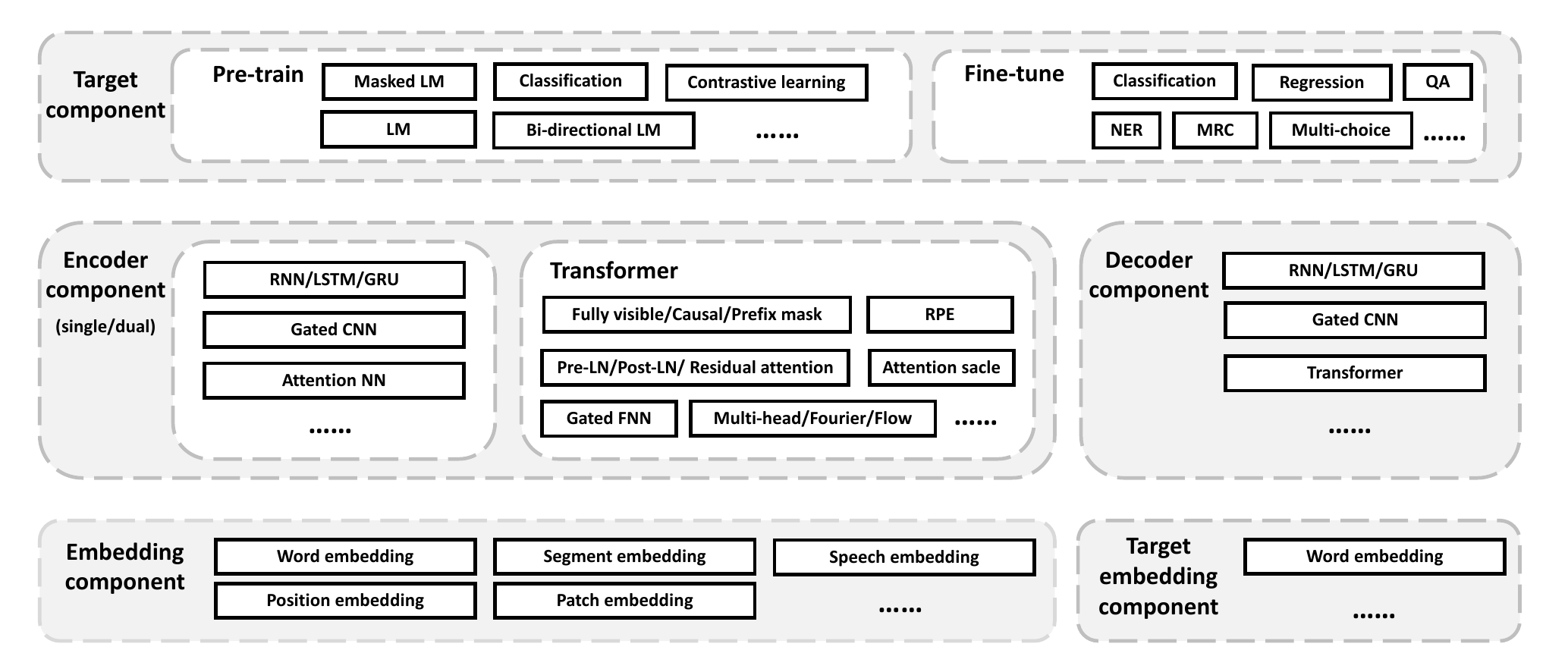}
\caption{The architecture of TencentPretrain. Pre-training models are implemented through module combination. TencentPretrain encourages reusing the existing modules and writing code at the module granularity.}
\end{figure*}

\section{Related Work}
\subsection{Pre-training models}
Pre-training models have been widely applied in text scenario. The success of pre-training is largely due to the powerful encoders for feature extraction (e.g., LSTM and Transformer), as well as the progress of pre-training target for learning knowledge from unsupervised corpus \cite{zhang2020pegasus,lewis2020bart,lan2019albert}. More recently, the text pre-training paradigm has been replicated in other modalities. For example, Transformer encoder (and its variants) has been widely used in vision \cite{dosovitskiy2020image}, audio \cite{gulati2020conformer,chen2022wavlm}, and vision-language tasks \cite{radford2021learning,kim2021vilt}. Regarding pre-training target component, text models have inspired models of other modalities. Mirroring the idea of masked language modeling (MLM), MAE \cite{he2022masked}, BEiT \cite{bao2021beit}, and SimMIM \cite{xie2022simmim} use masked image modeling (MIM) for self-supervised vision pre-training. Speech model Wav2vec2.0 \cite{baevski2020wav2vec} exploit negative sampling in pre-training target, which is previously used in word embedding \cite{mikolov2013distributed} and sentence prediction models \cite{logeswaran2018efficient, devlin2019bert, lan2019albert}.

In addition to the sharing of modules, several works have recently shown the feasibility of using the same pre-trained weight to handle different modalities simultaneously. For example, ERNIE-ViLG \cite{zhang2021ernie} and Talk2Face \cite{li2022talk2face} exploit prefix language model to achieve bi-directional text-and-image generation. PolyViT uses a single transformer model for image, video and audio classification \cite{likhosherstov2021polyvit}.

It can be seen that the trend towards homogeneity of pre-training models is becoming obvious, from sharing modules, to using the same network and parameters. This inspires us to build a unified framework that can implement various pre-training models efficiently.

\subsection{Toolkits with modular design}
Modular design regards a complex system as the combination of multiple modules, each of which can be independently modified and replaced. In the field of artificial intelligence, a typical work with modular design is Keras \cite{chollet2015keras}. The core data structure of Keras is layer. Keras allows building arbitrary graphs of layers to construct NN models. In the NLP field, modular toolkits are prevailing and they 
decompose models from different perspectives with different abstraction levels. For example, OpenNMT \cite{klein2017opennmt} is a modular toolkit designed for NMT. It builds an NMT model through the combination of encoder and decoder modules. Related NLP modular toolkits include OpenAttack (designed for text attack) \cite{zeng2021openattack}, Ngram2vec (designed for word embedding) \cite{zhao2017ngram2vec}, TextFlint (designed for robustness evaluation) \cite{wang2021textflint}, NeuralClassifier (designed for text classification) \cite{liu2019neuralclassifier}, and etc.

Inspired by the above-mentioned works, this paper proposes TencentPretrain, a modular designed toolkit for pre-training models of different modalities. Compared with Huggingface Transformers \cite{wolf2020transformers}, the most well-known pre-training toolkit, TencentPretrain provides additional abstractions on pre-training model implementations, splitting a complete model into multiple modules hierarchically. Pre-training weights between two toolkits can be switched easily. In fact, TencentPretrain can be regarded as the high-level encapsulation of Huggingface Transformers.

It is worth mentioning that TencentPretrain reuses part of the code in UER \cite{zhao2019uer}, which is published in 2019 and supports several text pre-training models. Compared with UER, TencentPretrain is improved in three aspects: 1) It supports the modular design within components, providing a more scalable manner to build pre-training models; 2) The \emph{\textbf{target embedding}} and \emph{\textbf{decoder}} components are introduced to support sequence generation; 3) In addition to text, TencentPretrain supports vision, audio, and cross-modal pre-training models. Currently, TencentPretrain supports around 30 pre-training models.

\section{Framework}

The current mainstream pre-training models are basically similar in structure. In the embedding component, the data is mapped into an embedding matrix. And then the matrix is passed through the encoder. Finally the target layer performs pre-training tasks according to the output of the encoder layer. If the pre-training task requires sequence generation, the decoder is inserted between the encoder and the target.

Figure 1 demonstrates the overall framework of TencentPretrain. It divides a pre-training model into five components, and various modules are provided in each component. In practice, a user firstly selects one or multiple modules from each component (modularization within component), and then combine modules from different components to build a pre-training model (modularization cross components). In the rest of this section, we respectively introduce the above five components and modules included in them.

\subsection{Embedding}
In the \emph{\textbf{embedding}} component, TencentPretrain converts text, image, and audio modal data into embedding matrix. The matrix holds the low-level features as the input to the encoder.

TencentPretrain also contains auxiliary embedding modules, e.g., position embedding and segment embedding. The embedding of pre-training model is usually obtained by the addition of multiple modules. As shown in Table 1 (Embedding column), the addition of word, position, and segment embeddings constitutes the embedding layer of BERT; the addition of patch and position embeddings constitutes the embedding layer of ViT. TencentPretrain supports hierarchical modular design, enabling users to freely combine modules within \emph{\textbf{embedding}} component to construct the desired embedding layer. This design greatly reduces code redundancy since different models often use similar, instead of identical combinations.

\subsection{Encoder}
TencentPretrain supports traditional encoders (e.g., LSTM and CNN) \cite{hochreiter1997long,kim2014convolutional}, as well as transformer and its variants (e.g., different normalization \cite{he2021realformer}, attention \cite{lee2021fnet}, masking strategies \cite{dong2019unified}). Users can construct customized transformer encoder by combining related options.

In addition, TencentPretrain supports dual-stream encoder, with which the users specify two encoder modules separately. Dual-stream encoder is usually used by models related with semantic search, such as text pair model SBERT \cite{reimers2019sentence} and text-image pair model CLIP \cite{radford2021learning}.

\subsection{Target embedding and decoder (optional)}
The pre-training tasks of some models involve sequence generation. These models require modules in \emph{\textbf{target embedding}} component and \emph{\textbf{decoder}} component. The modules in these two components are identical with the modules in \emph{\textbf{embedding}} component and  \emph{\textbf{encoder}} component respectively.

\subsection{Target}
The module in \emph{\textbf{target}} component receives high-level features obtained from encoder (or decoder) and then uses the features to perform pre-training tasks. Specifically, the target estimates gradients by objectives and updates the network weights. The target is of vital importance to the performance and has been extensively investigated in the pre-training field \cite{devlin2019bert,lan2019albert,sun2020ernie}. TencentPretrain supports comprehensive target modules, including language model \cite{radford2019language}, classification \cite{conneau2017supervised}, contrastive learning \cite{radford2021learning}, etc.

Sometimes pre-training models use multiple tasks, e.g., predicting word and sentence relationship simultaneously in BERT and ALBERT. And multi-task is especially common in cross-modal scenario \cite{kim2021vilt,lu2019vilbert,qi2020imagebert} since pre-training models have to deal with supervision signals from different modalities. The model can learn knowledge from different perspectives through multiple tasks. With the characteristic of hierarchical modular design, TencentPretrain facilitates the implementation of multi-task pre-training models. One can introduce multiple tasks by combining different modules in \emph{\textbf{target}} component. The pre-training task can be easily added, modified, and replaced.

\subsection{Downstream task fine-tuning}
TencentPretrain supports comprehensive downstream tasks, including classification, regression, sequence labeling, reading comprehension, question answering, automated speech recognition, etc. As shown in Figure 1, the downstream task model can be constructed by replacing the pre-training target with specific task. In evaluation section, we show the performances of TencentPretrain on a range of benchmarks.

\section{Usage}
This section provides examples of building pre-training models with TencentPretrain. The modular design enables the users to quickly build the pre-training model through the combination of modules. Modules used in pre-training models are specified in configuration files and the examples are shown as follows\footnote{Due to the page limit, we do not list entire configuration files. More details (e.g., Transformer encoder options) can be found in TencentPretrain project.}:




\lstset{escapeinside={<@}{@>}}
\definecolor{ao}{rgb}{0.0, 0.5, 0.0}
\definecolor{byzantine}{rgb}{0.5, 0.2, 0.64}

\begin{lstlisting}[basicstyle=\small]
<@\textcolor{ao}{\# BERT implementation}@>
{
  <@\textcolor{byzantine}{"embedding"}@>: ["word", "pos", "seg"],
  <@\textcolor{byzantine}{"encoder"}@>: "transformer",
  <@\textcolor{byzantine}{"target"}@>: ["mlm", "sp"]
}

<@\textcolor{ao}{\# T5 implementation}@>
{
  <@\textcolor{byzantine}{"embedding"}@>: ["word"]
  <@\textcolor{byzantine}{"encoder"}@>: "transformer"
  <@\textcolor{byzantine}{"tgt\_embedding"}@>: ["word"]
  <@\textcolor{byzantine}{"decoder"}@>: "transformer"
  <@\textcolor{byzantine}{"target"}@>: ["lm"]
}

<@\textcolor{ao}{\# ViLT implementation}@>
{
<@\textcolor{byzantine}{"embedding"}@>: ["patch_word", "pos", "seg"]
<@\textcolor{byzantine}{"encoder"}@>: "transformer"
<@\textcolor{byzantine}{"pooling"}@>: "first"
<@\textcolor{byzantine}{"target"}@>: ["cls", "mlm"]
}

<@\textcolor{ao}{\# CLIP implementation}@>
{
  <@\textcolor{byzantine}{"stream\_0"}@>:{
    <@\textcolor{byzantine}{"embedding"}@>: ["word", "pos"],
    <@\textcolor{byzantine}{"encoder"}@>: "transformer",
    <@\textcolor{byzantine}{"pooling"}@>: "first"
  }
  <@\textcolor{byzantine}{"stream\_1"}@>:{
    <@\textcolor{byzantine}{"embedding"}@>: ["patch", "pos"],
    <@\textcolor{byzantine}{"encoder"}@>: "transformer"
    <@\textcolor{byzantine}{"pooling"}@>: "first"
  }
  <@\textcolor{byzantine}{"target"}@>: ["clr"]
}
\end{lstlisting}

\begin{itemize}
\item BERT configuration file provides modules in \emph{\textbf{embedding}}, \emph{\textbf{encoder}}, and \emph{\textbf{target}} components. Since BERT has two pre-training tasks, its target is the combination of masked language model (mlm) and sentence prediction (sp).

\item T5 involves text generation. Its configuration file specifies modules used in \emph{\textbf{target embedding}} and \emph{\textbf{decoder}} components.

\item ViLT, an image-text pre-training model, is basically similar with text pre-training model BERT. The main difference is that an image-text embedding module is used in \emph{\textbf{embedding}} component.

\item CLIP is a dual-stream model. The modules in stream0 process text and the modules in stream1 process image. Contrastive learning (clr) module is used in  \emph{\textbf{target}} component.
\end{itemize}

If the desired pre-training model cannot be built by the combination of existing modules, TencentPretrain encourages users to develop a new module, and combine it with existing modules. We take the implementation of ASR model S2T \cite{wang2020fairseq} as an example. Most modules required by S2T are available and we only need to implement a new module, speech embedding, which greatly speeds up the implementation process.

TencentPretrain and Huggingface Transformers are interoperable. The conversion scripts are publicly available\footnote{https://github.com/Tencent/TencentPretrain/tree/main/scripts}, and the weights of different pre-training models can be converted between the two frameworks. In practical use, users are free to switch between these two frameworks.

With TencentPretrain, we build a pre-trained weight model zoo. Each pre-trained weight has two versions which can be loaded by either TencentPretrain or Huggingface Transformers. Currently, the TencentPretrain model zoo includes over 50 pre-trained weights. We provide pre-training data as well as training details, allowing users to reproduce results with less effort. The weights (pre-trained by TencentPretrain) are currently downloaded over 500 thousand times per month\footnote{https://huggingface.co/uer \\For Huggingface account, we inherit UER account instead of using TencentPretrain account.}. 

\begin{table}[h]
\centering
\small
\begin{tabular}{|c|c|c|c|}
\hline
\textbf{Model} & \textbf{HF} & \textbf{UER} & \textbf{TP} \\ \hline
Transformer&1135&749&795\\
\hline
+BERT&+822&\thead{+130\\\scriptsize{(+\textcolor{ao}{word\_pos\_seg},}\\\scriptsize{\textcolor{byzantine}{bert})}}&\thead{+149\\\scriptsize{(+\textcolor{ao}{pos},\textcolor{ao}{seg},}\\\scriptsize{\textcolor{byzantine}{mlm},\textcolor{byzantine}{sp})}}\\
\hline
+RoBERTa&+696&\thead{+92\\\scriptsize{(+\textcolor{ao}{word\_pos},\textcolor{byzantine}{mlm})}}&+0\\
\hline
+GPT-2&+688&+0&+0\\
\hline
+T5&+1008&+17\scriptsize{(+\textcolor{ao}{word})}&+0\\
\hline
+ViT&+493&-&+59\scriptsize{(+\textcolor{ao}{patch})}\\
\hline
+S2T&+824&-&+51\scriptsize{(+\textcolor{ao}{speech})}\\
\hline
+ViLT&+618&-&\thead{+15\\\scriptsize{(+\textcolor{ao}{word\_patch})}}\\
\hline
\end{tabular}
\caption{\label{few-tasks}
The number of code lines required for implementing a new pre-training model. The comment line in code is not counted. For UER and TencentPretrain, the added modules are also listed. \textcolor{ao}{Green} and \textcolor{byzantine}{violet} are used to denote embedding and target modules. Since UER does not support modularization within component, it has to introduce more modules (classes), e.g., word\_pos\_seg embedding and bert target, which are decomposed into multiple modules in TencentPretrain.
}
\vspace{-0.5cm}
\end{table}


\begin{table*}[!htbp]
\begin{small}
    \centering
    \begin{tabular}{|c|c|c|c|c|c|c|c|c|c|}

    \hline
    Model&MNLI&QNLI&QQP&RTE&SST&MRPC&CoLA&STS&AVG\\
    \hline
    BERT-base \scriptsize{(Ori.)} \cite{devlin2019bert}&83.9&90.7&90.7&65.7&92.3&88.9&56.5&88.6&82.2\\
    BERT-base \scriptsize{(DistilBERT)} \cite{sanh2019distilbert}&86.7&91.8&89.6&69.3&92.7&88.6&56.3&89.0&83.0\\
    BERT-base \scriptsize{(DynaBERT)} \cite{hou2020dynabert}&84.8&92.0&90.9&71.1&92.9&87.7&58.1&89.8&83.4\\
    BERT-base \scriptsize{(Metadistil)} \cite{zhou2022bert}&84.6&91.2&91.4&71.4&93.0&87.6&58.9&90.2&83.5\\
    BERT-base \scriptsize{(Ours)}&83.4&91.1&91.2&67.9&92.4&86.5&59.6&89.1&82.6\\
    \hline
    RoBERTa-large \scriptsize{(Ori.)} \cite{liu2019roberta}&90.2&94.7&92.2&86.6&96.4&90.9&68.0&92.4&88.9\\
    RoBERTa-large \scriptsize{(Ours)}&90.4&94.7&92.1&86.6&96.4&90.2&67.0&92.5&88.7\\
    \hline
    \end{tabular}

\end{small}
\caption{The comparison of TencentPretrain with other implementations on GLUE benchmark. We pre-train from scratch and then fine-tune on a range of datasets}
\end{table*}

\section{Evaluation}
This section evaluates TencentPretrain framework quantitatively. Firstly, we compare TencentPretrain with non-modular framework in terms of implementation cost. Then we show that TencentPretrain can reproduce the results of SOTA models on a range of benchmarks.

\subsection{Implementation cost}
The number of code lines is used to estimate the implementation cost. We only count the code lines in classes inheriting \emph{nn.Module}. We compare three frameworks, Huggingface Transformers (HF), UER, and TencentPretrain (TP). Huggingface Transformers exploits non-modular design. UER exploits semi-modular design, which doesn't support modularization within component.

When we continue to add new pre-training models (as shown in Table 2 from top to bottom), the number of code lines required by the TencentPretrain is less than the other two toolkits. Take RoBERTa as an example, TencentPretrain does not require any code work since it reuses modules for BERT. UER needs to add word\_pos module in \emph{\textbf{embedding}} component and mlm module in \emph{\textbf{target}} component. Huggingface Transformers builds a series of classes specific to RoBERTa, such as RoBERTaModel, RoBERTaEmbeddings, RoBERTaEncoder, RoBERTaPooler, which greatly increases the number of code lines. For other pre-training models, the conclusions are similar. The homogeneity among pre-training models makes the modular design much more advantageous.

In general, the code styles of Huggingface and TencentPretrain are inconsistent. Huggingface creates separate classes for each pre-training model, while TencentPretrain establishes generic modules that are independent of the specific model. Therefore, for most pre-training models, no additional code implementation is required in TencentPretrain.

\subsection{Reproducibility}
In this section, we follow the experimental settings of original papers. The scripts for running models on benchmarks are organized here\footnote{https://github.com/Tencent/TencentPretrain/wiki/\\Competition-solutions}, and users can easily reproduce the results in Table 3 and 4.

For text modality, we use GLUE benchmark to test TencentPretrain's performance. BERT-base and RoBERTa-large are used as test models. The results of BERT-base are listed in the first five rows in Table 3. As shown in AVG column, our result is 82.6, which falls into the range of 82.2-83.5 (the lowest and highest results reported by other papers). The average scores reported by DynaBERT and Metadistil are slightly higher than our result. One of the reasons is that development set of RTE only includes 277 instances, which leads to large fluctuations. The RTE results reported by DynaBERT and Metadistil are 3 point higher than our implementation. For RoBERTa-large, we can observe that our implementation results are close to the results reported by original RoBERTa paper.

Table 4 provides the results on vision and audio tasks. ViT \cite{dosovitskiy2020image} and BEiT \cite{bao2021beit} are used as test models for vision datasets. Top1 accuracy on vision datasets is reported. The original paper of BEiT only reports results on ImageNet. For audio dataset, we report the Automatic Speech Recognition (ASR) results on LibriSpeech with S2T \cite{wang2020fairseq}. Word Error Rate (WER) is shown in Table 4 (bottom). We can observe that the results of TencentPretrain are close to the results reported by original papers.

\begin{table}[h]
\centering
\small
\begin{tabular}{|c|c|c|c|}
\hline
Model & CIFAR10 & CIFAR100 & \thead{ImageNet\\1000} \\ \hline
ViT-base&98.95&91.67&83.97\\
ViT-base\scriptsize{(Ours)}&98.73&92.12&83.97\\
BEiT-large&-&-&87.30\\
BEiT-large\scriptsize{(Ours)}&-&-&87.24\\
\hline
\end{tabular}

\vspace{0.5cm}

\begin{tabular}{|c|c|c|c|c|}
\hline
Model & devclean & devother & testclean & testother \\ \hline
\footnotesize{S2T}&3.8&8.9&4.4&9.0\\
\footnotesize{S2T\scriptsize{(Ours)}}&3.8&9.2&4.1&9.0\\
\hline
\end{tabular}

\caption{The comparison of TencentPretrain with original implementations on datasets of vision and audio modalities.}
\vspace{-0.3cm}
\end{table}

\section{Conclusion}
This paper presents TencentPretrain, a pre-training toolkit characterized by modular design and multi-modal support. In TencentPretrain, pre-training models of different modalities are regarded as the combination of multiple modules, which is easy to configure and extensible. Furthermore, we quantitatively demonstrate that TencentPretrain facilitates the users to reuse existing modules and decreases the cost of model development. At last, we test TencentPretrain on a range of datasets and show that it can reproduce the SOTA results.

\section{Limitations}

Although the TencentPretrain pre-training framework has integrated optimization libraries like Deepspeed and Apex, it still lacks support for other components such as Megatron.
In the future, we will provide more parallelism modes to achieve efficient training of large-scale language models (LLM).

\section*{Acknowledgements}
The work was supported by the National Natural Science Foundation of China under Grant No. 62072458. We are grateful to Lusheng Zhang, Xiayu Li, Jian Kang, Jinwen Luo, Weidong Guo, Jiachi Liu, Jianwei Cui, Dongxiao Huang, Xingyu Bai, Yang Xu, Huanqin Wu, Tongwen Huang, Peng Meng, Yanming Xu, Chunquan Chen, Xuefeng Yang, Qi Ju for code contribution. We also received helpful ideas and feedback from members of the TencentNLP Oteam and Institute of Computer Vision, Shenzhen University.

\bibliography{anthology,custom}
\bibliographystyle{acl_natbib}

\end{document}